\documentclass{article}

     \usepackage[preprint,nonatbib]{neurips_2020}

\usepackage[utf8]{inputenc} 
\usepackage[T1]{fontenc}    
\usepackage{hyperref}       
\usepackage{url}            
\usepackage{booktabs}       
\usepackage{amsfonts, amsmath}       
\usepackage{nicefrac}       
\usepackage{microtype}      
\usepackage{wrapfig}
\usepackage{graphicx}
\usepackage{bm}

\newcommand{\set}[1]{\left\{#1\right\}}

\newcommand{\RR}{\mathbb{R}}

\newcommand{\HH}{\mathbb{H}}
\newcommand{\bbS}{\mathbb{S}}
\newcommand{\II}{\mathbb{I}}

\newcommand{\cB}{\mathcal{B}}

\newcommand{\cL}{\mathcal{L}}

\newcommand{\HP}{\mathrm{HP}}
\newcommand{\dist}{\mathrm{d}}
\DeclareMathOperator{\arcosh}{arcosh}
\DeclareMathOperator{\argmin}{arg\,min}
\DeclareMathOperator{\argmax}{arg\,max}

\title{A Theory of Hyperbolic Prototype Learning}

\author{%
  Martin Keller-Ressel\\
  Department of Mathematics\\
  TU Dresden\\
  Germany\\
  \texttt{martin.keller-ressel@tu-dresden.de} \\
}

\begin{document}

\maketitle

\begin{abstract}
We introduce Hyperbolic Prototype Learning, a type of supervised learning, where class labels are represented by ideal points (points at infinity) in hyperbolic space. Learning is achieved by minimizing the `penalized Busemann loss', a new loss function based on the Busemann function of hyperbolic geometry. We discuss several theoretical features of this setup. In particular, Hyperbolic Prototype Learning becomes equivalent to logistic regression in the one-dimensional case.
\end{abstract}

\section{Introduction}

Prototype learning is a type of supervised learning, where class labels are represented by `prototypes', that is, by points in a suitably chosen output space. Instead of minimizing a general loss function, the learning model is trained by minimizing the distance of its output to the prototype of the true class label. In particular for multi-class categorization, prototype learning is a viable alternative to more common representations of class labels, such as `one-hot encoding' or word2vec, see \cite{mettes2019hyperspherical}.\\
The crucial ingredients of a prototype learner are the choice of its output space (including its metric structure) and a method to embed prototypes into the output space. In \cite{mettes2019hyperspherical} \textit{hyperspherical prototype learning} is proposed, in which the output space is given by the $d$-dimensional sphere $\bbS^d$ and improvements over one-hot-encoding and word2vec are shown. Here, we formulate a theory of hyperbolic learning, in which the output space is given by $d$-dimensional \textit{hyperbolic space} $\HH^d$ and prototypes are represented by \textit{ideal points} (points at infinity) of $\HH^d$. Instead of hyperbolic distance, we propose to use the (penalized) \textit{Busemann function} which can be interpreted as a `distance to infinity'  and can be meaningfully applied to ideal points. We show that in the one-dimensional case, hyperbolic prototype learning with penalized Busemann loss and a linear base learner is equivalent to logistic regression, or -- coupled with a general neural network -- equivalent to cross-entropy-loss combined with a logistic output function. 

\section{Hyperbolic Geometry}

\subsection{The Poincar\'e ball model of hyperbolic space}

\begin{wrapfigure}{r}{0.4\textwidth}
\begin{center}
\includegraphics[width=0.35\textwidth]{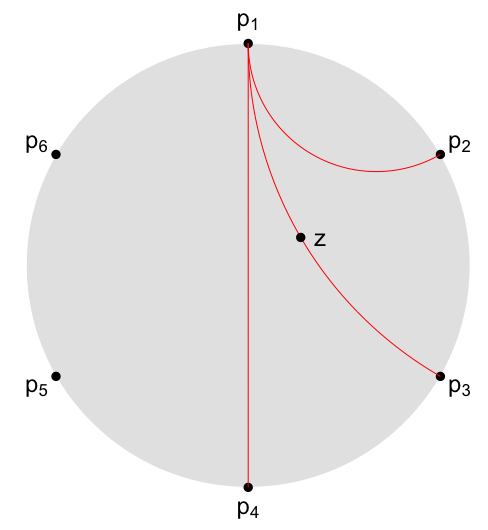}
\caption{The Poincar\'e disc with six ideal points $p_1, \dotsc, p_6$ and an `ordinary' point $z$.  Three geodesics are shown in red. The point $z$ is at infinite hyperbolic distance, but at finite `Busemann distance', from all ideal points $p_i$.\label{fig:schematic}}
\end{center}
\end{wrapfigure}

In the Poincar\'e ball model, $d$-dimensional hyperbolic space is represented by the open unit ball
\[\HH^d = \set{z \in \RR^d: z_1^2 + \dotsm  + z_d^2 < 1}.\]
We parameterize $\HH^d$ by hyperbolic polar (HP) coordinates $(r,u)_\HP$, consisting of a unit vector $u \in \bbS^d$ and the hyperbolic radius $r \in [0,\infty)$, such that 
\[z = \tanh(r/2)\,u,\]
i.e., $|z| = \tanh(r/2)$ is the Euclidean norm of $z$. Equipped with the \textit{hyperbolic distance}
\[\dist_H((r_1,u_1)_\HP,(r_2,u_2)_\HP) = \arcosh\Big(\cosh(r_1)\cosh(r_2)  - \sinh(r_1)\sinh(r_2)\, u_1 \cdot u_2\Big),\]
$\HH^d$ becomes a metric space, cf. \cite{ratcliffe2006foundations, cannon1997hyperbolic}. In two dimensions, $\HH^2$ is called the Poincar\'e disc and is convenient to visualize properties of hyperbolic geometry\footnote{An artistic rendition of the Poincar\'e disc is given by the woodcuts `Circle Limit I-IV' by the Dutch artist M.C. Escher; see \url{https://www.wikiart.org/en/m-c-escher}.}, see Figure~\ref{fig:schematic}. For $d=2$, the direction vector $u$ can be replaced by its angle $\theta \in [0,2\pi)$ such that the hyperbolic distance becomes
\[\dist_H((r_1,\theta_1)_\HP,(r_2,\theta_2)_\HP) = \arcosh\Big(\cosh(r_1)\cosh(r_2)  - \sinh(r_1)\sinh(r_2) \cos(\theta_1 - \theta_2)\Big).\]
Endowed with the metric tensor
\[ds^2 = \frac{4|dz|^2}{(1 - |z|^2)^2}\]
$\HH^d$ becomes a Riemannian manifold and $\dist_H$ is precisely the corresponding Riemannian distance (see \cite{ratcliffe2006foundations}). Hence, all concepts from differential geometry, such as tangent space, geodesics, (sectional) curvature, and exponential maps have direct interpretations in the context of hyperbolic space. Here, we will only make use of the exponential map\footnote{Roughly speaking, the exponential map $\exp_p(y)$ returns the result of following a geodesic from $p$ with speed $|y|$ and in direction $y / |y|$.} at the origin of $\HH^d$, which is given (in Euclidean and in hyperbolic polar coordinates) by
\[\exp_0: \quad \RR^d \to \HH^d, \quad  y \mapsto \tanh(|y|/2) \frac{y}{|y|}  = \left(|y|, \frac{y}{|y|}\right)_\mathrm{HP}.\]
\subsection{Ideal points and the Busemann function}
Ideal points in hyperbolic geometry represent points at infinity. In the Poincar\'e model, the ideal points form the boundary of the unit ball, i.e., they are given by   
\[\II^d = \set{z \in \RR^d: z_1^2 + \dotsc z_d^2 = 1}\]
and each ideal point is naturally associated with a unit vector $p \in \bbS^d$. In hyperbolic coordinates this corresponds to a point with infinite radial coordinate, i.e. we can write $p = (\infty, p)_\HP$.\\
As outlined above, our goal is to represent class prototypes by ideal points. This raises the problem that ideal points are at infinite distance from all other points in $\HH^d$ and hence that the hyperbolic distance cannot be used as a loss function for prototype learning. This problem can be avoided by replacing hyperbolic distance by the \textit{Busemann function}. The Busemann function, originally introduced in \cite{busemann1955geometry} (see also Def.~II.8.17 in \cite{bridson2013metric}) can be considered a `distance to infinity' and is defined (in any metric space) as follows: Let $p$ be an ideal point and $\gamma_p$ a geodesic ray (parameterized by arc length) tending to $p$. Then the Busemann function with respect to $p$ is defined for $z \in \HH^d$ as
\[b_p(z) = \lim_{t \to \infty} (d_H(\gamma_p(t),z) - t).\]
In the Poincar\'e model this limit can be explicitly calculated and the Busemann function is given by
\[b_p((r,u)_\HP) = \log \Big(\cosh(r) - \sinh(r)\,u \cdot p\Big),\]
or, alternatively, in Euclidean coordinates by
\[b_p(z) = \log\left(\frac{|p - z|^2}{1 -|z|^2}\right).\]
In the two-dimensional case, replacing $u$ and $p$ by the angles $\theta$ and $\xi$, we obtain
\[b_\xi((r,\theta)_\HP) = \log \Big(\cosh(r) - \sinh(r) \cos(\xi - \theta)\Big).\]

\section{Hyperbolic Prototype Learning}
\subsection{Hyperbolic prototypes and the penalized Busemann loss}
A Hyperbolic Prototype Learner consists of 
\begin{itemize}
    \item A base learner $\cB(x,w)$ which, given trainable weights $w$, maps an input $x \in \RR^n$ to an output $y \in \RR^d$. Examples are linear regression ($\cB(x,w) = w^\top x + w_0$) or  multi-layer feed-forward networks. 
    \item A representation of class labels $1, \dotsc, K$ as prototypes $P = \set{p_1, \dotsc p_K}$ in the set of ideal points $\II^d$.
\end{itemize}
For training, the base learner $\cB$ is concatenated with the exponential map\footnote{See also \cite{chami2019hyperbolic}, where the exponential map of $\HH^d$ is used as a transfer function between the layers of a graph convolutional network.} $\exp_0$, such that each input $x$ is mapped to a point
\[z = \exp_0(\cB(x,w))\]
in $\HH^d$. To evaluate the output $z \in \HH^d$ against the prototype $p_j$ of the true class label $j \in [K]$, we propose to use the \textit{penalized Busemann (peBu) loss}
\[l(z;p) = b_{p}(z) - \log(1 - z^2) = 2 \log \left(\frac{|p-z|}{1 - |z|^2}\right),\]
or, alternatively in hyperbolic polar coordinates
\[l((r,u)_\HP; p) = \log \Big(\cosh(r) - \sinh(r)\,u \cdot p\Big) + \log\Big(\cosh(r) + 1\Big).\]
The role of $b_p(z)$ is to steer $z$ towards the prototype $p$, while the penalty term penalizes `overconfidence', i.e., values of $z$ close to the ideal boundary of $\HH^d$. The exact form of the penalty can be motivated from the fact that the peBu-loss becomes identical (up to scaling) to cross-entropy-loss in dimension $d=1$; see below. 
Given a training sample $(x_i,p_i)_{i=1}^N$ of inputs $x_i$ with class labels represented by prototypes $p_i$, the Hyperbolic Prototype Learner is trained by minimizing the sample peBu-loss of the embedded base learner output, i.e. by minimizing
\[\cL(w) = \frac{1}{N} \sum_{i=1}^N l\Big(\exp_0(\cB(x_i,w));p_i\Big).\]

\subsection{Prediction from a Hyperbolic Prototype Learner}
For prediction, we propose the same procedure as in \cite{mettes2019hyperspherical}: For a given input $x$, the class label of the closest prototype to $z = \exp_0(\cB(x,w))$ is returned, i.e. the predicted prototype is
\[p_* = \argmin_{p \in P} l(z;p).\]
As the penalty term $-\log(1 - |z|^2)$ is the same for all prototypes, we can equivalently minimize the Busemann function $b_p(z)$ directly over prototypes. Moreover, as $b_p(z)$ is (for given $z$) a decreasing function of the cosine similarity $\tfrac{z}{|z|} \cdot p$, prediction is equivalent to maximizing the cosine similarity between $\tfrac{z}{|z|}$ and $p$, that is
\[p_* = \argmax_{p \in P} \frac{z}{|z|} \cdot p.\]
Note that this is exactly the same prediction procedure as in hyperspherical embedding (see eq.~(3) in \cite{mettes2019hyperspherical}). This also implies that each output $z = (r,u)_\HP$ can be separated into the directional coordinate $u$, which measures the \textit{similarity} to a given prototype $p$, and the radial coordinate $r$, which represents the \textit{confidence} of this assessment and which can be interpreted analogous to the `log-odds' in logistic regression.\footnote{Interestingly, a similar decomposition of hyperbolic coordinates into a `similarity' and a `popularity' component is at the heart of the influential network growth model (`PSO-model') of \cite{papadopoulos2012popularity}.}


\subsection{Gradients of the penalized Busemann loss}
The gradients of the penalized Busemann loss with respect to $(r,u)_\HP$ can be easily calculated and we obtain
\begin{align*}
\partial_r\,l((r,u)_\HP;p) &= - \frac{p \cdot u - \tanh(r)}{1 - \tanh(r) \, p \cdot u} + \tanh(r/2)\\
\nabla_u\,l((r,u)_\HP;p) &= - p\,\frac{\tanh(r)}{1 - \tanh(r) \, p \cdot u}.
\end{align*}
Also the gradient of $l(\exp_0(y);p)$ with respect to $y$, which is need for backpropagation to the base learner $\cB$, can be calculated with some effort and is given by
\begin{multline*}
\nabla_y\,l(\exp_0(y);p) = (y - p) \frac{\tanh(|y|}{|y| - \tanh(|y|) \, p \cdot y} + \bm{1}\, p \cdot y \frac{\tanh(|y|)/|y| - 1}{|y| - \tanh(|y|) \, p \cdot y} + \tanh(|y|/2),
\end{multline*}
where $\bm{1}$ is a vector of $d$ ones.\\
 
\begin{figure}[!h]
\begin{center}
\includegraphics[width=0.4\textwidth]{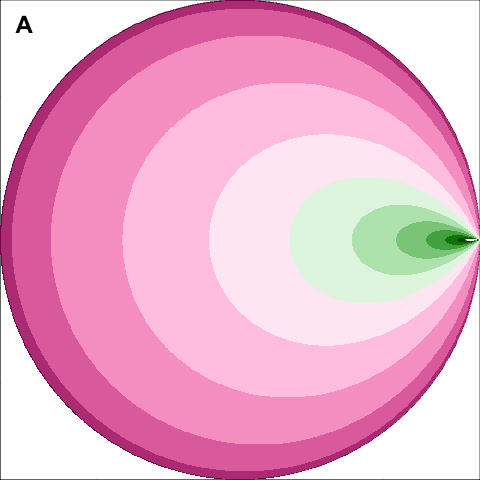}
\hspace{2em}
\includegraphics[width=0.4\textwidth]{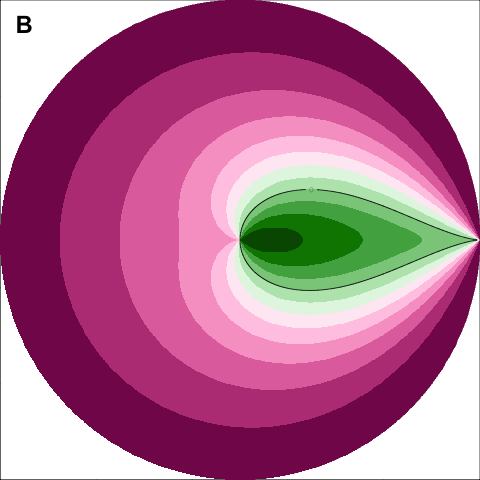}
\caption{Penalized Busemann loss (A) and its radial gradient (B) for a prototype located at the ideal point $(1,0)$. \label{fig:loss}}
\end{center}
\end{figure}

\subsection{Equivalence to logistic regression for $d=1$}
We proceed to show that in the one-dimensional case ($d=1$) hyperbolic prototype learning with linear base learner and penalized Busemann loss is equivalent to logistic regression. More generally, the case $d=1$ is equivalent to a concatenation of the base learner $\cB$ with a logistic output function and cross-entropy-loss.\\
In one dimension, hyperbolic space $\HH^1$ becomes the interval $(-1,1)$. The set of ideal points $\II^1$ consists of the two points $\pm 1$. Thus, only two prototypes $p_+, p_-$ can be embedded at the points $z = \pm 1$, which corresponds to a binary classification task. In binary classification, class labels are more commonly identified with $0/1$, and hence we introduce another simple (linear) change of coordinates to 
\[z' = \frac{z+1}{2}.\]
Under this change of coordinates $\HH^1$ becomes the unit interval $(0,1)$ and the prototypes $p'$ (the ideal points) become the endpoints $0$ and $1$. The exponential map, i.e., the embedding of the base learner output $y = w^\top x + w_0$ into $\HH^1$, maps $y \in \RR$ to 
\[z' = \exp_0(y) = \frac{\tanh(y/2) + 1}{2} = \frac{1}{1 + e^{-y}},\]
which is the logistic function. In $z'$-coordinates, the penalized Busemann loss becomes
\[l(z';p') = 2 \log\Big(\frac{|p' - z'|}{2z'(1 - z')}\Big) =  - 2p' \log \Big(2(1-z'))  - 2(1 - p') \log(2z')\Big).\]
Using the cross entropy
\[h(z';p') = - p' \log(z') - (1 - p') \log(1-z'), \]
we can write the peBu-loss as
\[l(z';p') = 2 h(z';p')  - 2 \log(2),\]
a scaled and shifted transformation of cross-entropy. We conclude that minimizing the peBu-loss is equivalent to minimizing cross-entropy in the one-dimensional case. 

\subsection{Embedding of prototypes}
The set of ideal points of $\HH^d$ is homeomorphic to the hypersphere $\bbS^d$. Thus, any of the methods proposed for prototype embedding into $\bbS^d$ in \cite{mettes2019hyperspherical} can also be used to embed prototypes into $\II^d$. In particular
\begin{itemize}
    \item in dimension $d=2$ prototypes can be placed uniformly onto the unit sphere $\bbS^2$
    \item in dimension $d > 2$ prototypes can be placed by minimizing their mutual cosine similarities, as proposed in \cite{mettes2019hyperspherical}. Since the peBu-loss is a decreasing function of the cosine similarity between directional coordinate and prototype, the arguments outlined in \cite{mettes2019hyperspherical} for hyperspherical prototypes equally apply to hyperbolic prototypes.
    \item A hyperbolic embedding method, such as \texttt{hydra} (see \cite{keller2020hydra}), can be used to embed prototypes into $\HH^d$. This results in coordinates $\tilde p_j = (r_j,u_j)_\HP$ for each class label  $j \in [K]$, which can be projected to the ideal boundary by setting $p_j = (\infty,u_j)_\HP$.
\end{itemize}
\begin{wrapfigure}[14]{r}{0.4\textwidth}
\vspace{-3.5em}
\begin{center}
\includegraphics[width=0.35\textwidth]{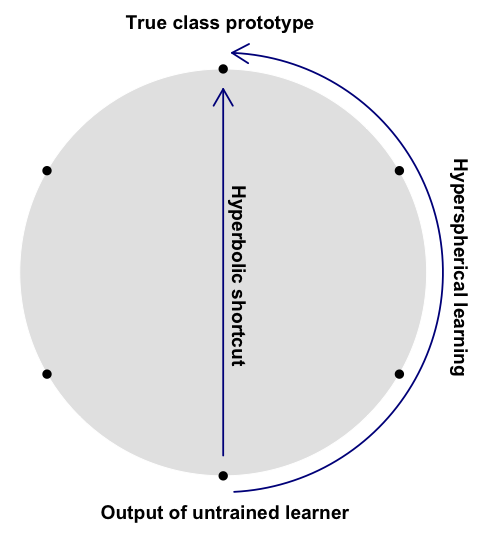}
\caption{The `hyperbolic shortcut' \label{fig:shortcut}}
\end{center}
\end{wrapfigure}
\subsection{The `hyperbolic shortcut'}
Finally, we present a heuristic argument, why we expect hyperbolic prototype learning to be more efficient than hyperspherical prototype learning. Suppose that an untrained learning model gives an output which is `as wrong as possible', i.e., opposite from the correct prototype (see Figure~\ref{fig:shortcut}). To update its output toward the true prototype, the hyperspherical learner has to `walk along the sphere', passing through several other incorrect prototypes to reach the correct prototype.\footnote{This problem is somewhat alleviated in higher dimension.} The hyperbolic learner, on the other hand, can take a shortcut and `cross through the disc' without passing through other incorrect prototypes.

\section{Summary} We have proposed Hyperbolic Prototype Learning, a type of supervised learning, which uses hyperbolic space as an output space. Learning is achieved by minimizing the penalized Busemann loss function, which is given by the Busemann function of hyperbolic geometry, enhanced by a penalty term. As an interesting property, we have shown that Hyperbolic Prototype Learning becomes equivalent to logistic regression in dimension one, when a linear base learner is used. It remains to be seen whether the performance of a practical implementation of Hyperbolic Prototype Learning can live up to its attractive theoretical features.

\small
\bibliographystyle{plain}
\bibliography{references}

\end{document}